\documentclass[conference]{IEEEtran}
\usepackage{cite}
\usepackage{amsmath}
\usepackage{graphicx}
\usepackage{adjustbox}
\usepackage{array}
\usepackage{filecontents}

\begin{document}

%Here goes the title

\title{Bench-Marking Information Extraction in Semi-Structured Historical Handwritten Records}

%Authors List

\author
{ \IEEEauthorblockN{\textbf{Animesh Prasad$^\dagger$, Herv\'e D\'ejean$^\ddagger$, Jean-Luc Meunier$^\ddagger$} \\ \textbf{Max Weidemann$^\ast$, Johannes Michael$^\ast$, Gundram Leifert$^\ast$}}

 \IEEEauthorblockA{$^\dagger$National University of Singapore, $^\ddagger$Naver Labs, $^\ast$University of Rostock\\
 a0123877@u.nus.edu, \{herve.dejean,
 jeanluc.meunier\}@naverlabs.com \\
 \{max.weidemann,  johannes.michael,
gundram.leifert\}@uni-rostock.de}
}
\maketitle

%Main body starts

\begin{abstract}
In this report, we present our findings from benchmarking experiments for information extraction on historical handwritten marriage records Esposalles from IEHHR - ICDAR 2017 robust reading competition. 
The information extraction is modeled as semantic labeling of the sequence across 2 set of labels. This can be achieved by sequentially or jointly applying handwritten text recognition (HTR) and named entity recognition (NER). We deploy a pipeline approach where first we use state-of-the-art HTR and use its output as input for NER. We show that given low resource setup and simple structure of the records, high performance of HTR ensures overall high performance. We explore the various configurations of conditional random fields and neural networks to benchmark NER on given certain noisy input. The best model on 10-fold cross-validation as well as blind test data uses n-gram features with bidirectional long short-term memory.
\end{abstract}

\begin {IEEEkeywords}
Handwritten Text Recognition, Information Extraction, Named Entity Recognition, CRF, LSTM
\end{IEEEkeywords}

\section{Introduction}
\label{intro}
Lately, the interest of the document image analysis community in document understanding, information extraction and semantic categorization is waking in order to make digital search and access ubiquitous  for archival documents
 An example of such information extraction is NER in demographic documents. Information may contain people's names, birthplaces, occupations, etc in some structured (like tables) or semi-structured (like records or entries) format. 
 Extracting targeted information from various such sources and converting it in structured repositories (like databases) can allow the access to their contents and envision innovative services based in genealogical, social or demographic searches.
 
Recently there are various techniques based on hidden Markov model (HMM), conditional random fields (CRF), long short term memory (LSTM) and convolutional neural network (CNN) proposed for handwritten text recognition (HTR) as well as for various natural language tagging tasks.
 
The parameters of the IEHHR  i.e. amount of data, target information, complexity of the task, performance of HTR etc. all conform to general use cases of historical information extraction. 
In this report we try to benchmark such techniques on basic information extraction task on semi-structured records modeled as tagging under given high performance of HTR. We hope by testing with various configurations of state-of-the-art tagging techniques we would be able to identify strong baselines for NER on noisy text generated from some off-the-shelf HTR.

\textbf{Dataset.}  The competition used 125 pages of the Esposalles database \cite{romero2013esposalles}, a marriage license book conserved at the archives of the Cathedral of Barcelona.
%Animesh: @Herve Please verify the size of the dataset
%Hervé: their site is down...
The corpus is written in old Catalan by only one writer in the 17th century. Each marriage record contains information about the husband’s occupation, place of origin, husband’s and wife’s former marital status, parents’ occupation, place of residence, geographical origin, etc. The structure of the marriage record tends to follow a regular expression (with some exceptions):

\setlength{\fboxrule}{0pt}
\vspace{0.2cm}
\noindent\fbox{%
    \parbox{8.5cm}{%
    \small
\textit{$<$husband$>$} fille de \textit{$<$husband's father$>$} y \textit{$<$husband's mother$>$} ab \textit{$<$wife$>$} fille de \textit{$<$wife's father$>$} y \textit{$<$wife's mother$>$}

\vspace{0.2cm}

\textit{$<$husband$>$} fille de \textit{$<$husband's father$>$} y \textit{$<$husband's mother$>$} ab \textit{$<$wife$>$} viusa \textit{$<$wife's former husband$>$}
 }%
}
 
 \vspace{0.2cm}
 
\begin{figure}[ht]
   \centering
   \includegraphics[scale=.33]{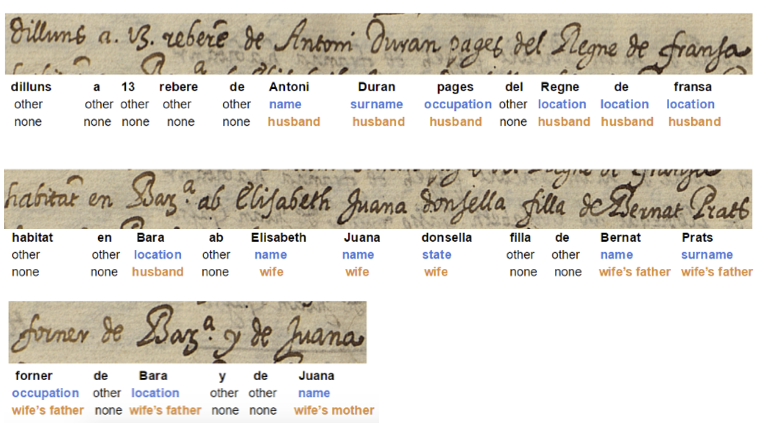}
    \caption{An example of a record}
    \label{fig:record}
\end{figure}

\textbf{Tasks.}
The objective is to extract information from the records in simplified predefined semantic classes.  The marriage records are manually annotated at token, lines and the level of the record with semantic annotations for each token. 
The training and test sets are composed of:
\begin{itemize}
    \item  Training set: 100 pages, 968 marriage records.
    \item  Test set: 25 pages, 253 marriage records. 
    %Animesh: @Herve Please verify the size of dataset
 \end{itemize} 
    For each marriage record (Fig. \ref{fig:record}), we use:
 \begin{itemize}
    \item  Images of segmented text lines.
    \item  Text files with the corresponding transcription.
    \item  Text files with the corresponding categories: name, surname, occupation, location, and state.
    \item  Text files with the corresponding person: husband, husband’s father, husband’s mother, wife, wife’s father, wife’s mother and other-person.
\end{itemize}

%Animesh: @Herve Image used from original competition site, allowed?
%Hervé: yes

\begin{figure}[ht]
   \centering
   \includegraphics[scale=.25]{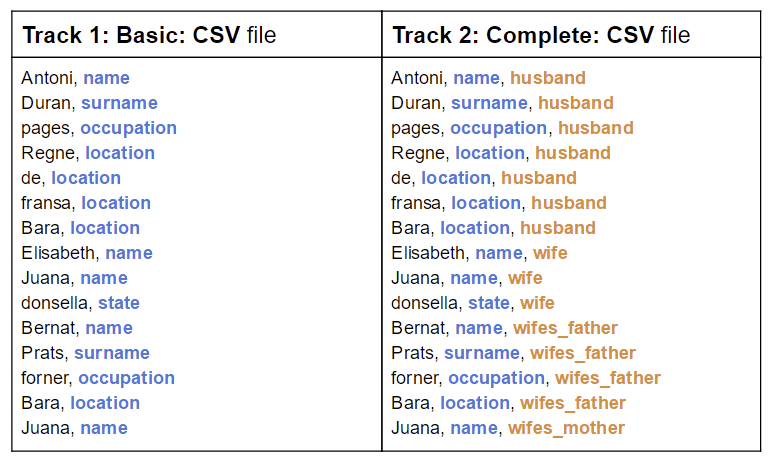}
    \caption{Sample Evaluation Scheme}
    \label{fig:record_label_scheme}
\end{figure}

 For evaluation on blind test data, the CSV file with the transcription of the relevant words (named entities) and their semantic category is generated for each record.  This represents an evaluation metric to simulate the filling in of a knowledge base.
 An example of labeled record (training sample) and it's named entity (expected output) is shown in Fig. \ref{fig:record} and Fig. \ref{fig:record_label_scheme} respectfully.

%Animesh: @Herve Image used from original competition site, allowed?
%Hervé: yes

The evaluation is performed on 2 tracks defined as follows:
\begin{itemize}
\item \textbf{Basic} The CSV must contain the transcription and the categories (name, surname, occupation, etc.) (say \textit{Task 1})
\item \textbf{Complete} The CSV must contain the transcription, the categories and the person (husband, wife, wife’s father, etc.) (say \textit{Task 1} and \textit{Task 2} respectively)
\end{itemize}

\section{Related Work}
\label{section2}
For handwritten text recognition wide variety of neural architectures like CNN, BLSTM etc. have been shown to work better than traditional feature-based HMM, CRF architectures \cite{strauss2014citlab}.
Further, various techniques have been devised for text segmentation and skew normalization for preprocessing images to enhance HTR. Together with such signal processing techniques, deep architectures and recent language modeling advancements have resulted into reasonable performance for many images to text systems.    

In the natural language processing (NLP) domain, deep neural network based models obtained comparable results to state-of-the-art, human-engineered feature-based models on part-of-speech (POS) tagging, chunking, NER, and semantic role labeling (SRL) \cite{Collobert:2011:NLP:1953048.2078186}.
Such tasks are all sequence labeling tasks, in which the system assigns labels to variable-length sequences.
For NLP tasks, task-specific features while obtaining good results have been often facilitated by pretrained word embedding.
Specifically for NER various end-to-end architectures using word embedding (optionally with character-based word embedding) features learned via neural models  (BLSTM, BLSTM-CNN \cite{lample2016neural}) together with sequential decoding of CRF is used. 

Information extraction on historical handwritten text has gained traction in light of efforts for digitizing old documents and making it accessible to search. There has been prior work on keyword spotting \cite{frinken2012novel}, database completion \cite{dejean2015extracting} etc. 
Such work either follow a pipeline based approach as in \cite{meunier2018document} where first we localize the semantic data of interest (like address in a tabular data) and then use Regular Expressions (RegEx) on the localized text to extract the data, or jointly doing localization and extraction \cite{dejean2015extracting}.
%Animesh: @Herve  verify this last paragraph on the usage of NLP in DU

\textbf{State-of-the-art.} The IEHHR  saw techniques spanning both the aforementioned approaches \cite{fornes2017icdar2017}.
 Hitsz-ICRC-1 and Hitsz-ICRC-2 use a pipeline based approach where first they use a CNN or ResNet respectively to extract the character sequence out of the image followed by bi-gram frequency based postprocessing for improving recognition performance. In the second stage they use CRF sequence
tagging method via the CRF++ over the output of the HTR. These model use word level image segments to train HTR.
 The  CITlab-ARGUS-1, CITlab-ARGUS-2, and CITlab-ARGUS-3 use CNN-RNN model trained with CTC along with data augmentation. However, they directly use the character probability matrix generated by the CTC along with the location information to identify the region of interest and apply RegEx directly on the matrix. These model use line level image segments to train HTR.
 Further, recently joint modeling techniques were proposed \cite{carbonell2018joint} performing both the tasks in an end-to-end fashion. Being a joint model they append the line level images to get the overall record image. 

The potential shortcoming of both the models is that they use simple CRF or RegEx based approach for NER based which might not be robust to HTR sensitivities. Further, the end-to-end model might not be fit for the task given the amount of data. A suitable compromise isn't explored between the deep learning based modeling and robust feature extraction.

\section{Method}
\label{section3}
Our method is divided into two parts, HTR and NER respectively. First, we explain how the HTR output is generated from the images and then we go through the various tagging models experimented for the NER part. 
%Animesh: @Herve  verify whole following subsection is copied directly from CITlab's arxiv paper, alright?
%Hervé @animesh: mention the source (arxiv paper)
 
\subsubsection{\textbf{Handwritten Text Recognition }}
In the HTR stage \cite{strauss2018system}, we present a novel, segmentation-free, word-wise character recognition method without any external linguistic knowledge. In this method, the position information of each character is converted into a vector. A kind of uni-gram model is then constructed and integrated into the residual neural network for training. The whole process of character recognition consists of following steps:
 \begin{itemize}
     \item  \textbf{Data Preprocessing.} Given the line polygon, we apply certain standard preprocessing routines, i.e. -- image normalization: contrast enhancement (no binarization), size; -- writing normalization: line bends, line skew, script slant. Then,  images are further unified by writing normalization: The writing’s main body is placed in the center part of an image of fixed 96px height. While the length-height ratio of the main body stays untouched, the ascenders and descenders are squashed to focus the network’s attention on the more informative main body.
     
     \item \textbf{Model Training} and  \textbf{Generating Output.} The preprocessed images are fed into a neural network of the architecture as described next. The implementation is based on TensorFlow \cite{abadi2016tensorflow}. The 3 convolutional layers additionally apply batch normalization \cite{ioffe2015batch} before and local response normalization \cite{krizhevsky2012imagenet} after applying the ReLU activation function and dropout of 0.5. The last layer is the fully-connected layer and contains  62  neurons. One of these neurons represents a garbage label (not-a-character or NaC in the following) and the others correspond to the 61 characters appearing in the ground truth.
     Overall network configuration is  conv, conv, BLSTM, conv, BLSTM, dense (8, 32, 256, 64, 512, 62 neurons respectively).
     The loss function is the typical CTC loss \cite{graves2006connectionist}. The network is trained 150 epochs by RMSProp where one epoch contains 4096 randomly sampled line images. The initial learning rate is 0.002 and decayed after every third epoch by a factor of 0.95. The output of the last layer is softmax transformed such that the output of the neural network is a matrix. The final output from this part is best effort character decoding of each line image. 

  \end{itemize}
  
  Our method result into high-accuracy character recognition with character error rate (CER) of only around 5\%.

\newcolumntype{C}[2]{%
    >{\adjustbox{angle=#1,lap=\width-(#2)}\bgroup}%
    c%
    <{\egroup}%
}
\newcommand*\rot{\multicolumn{1}{C{90}{1em}|}}

{\setlength{\tabcolsep}{3.5pt}
\renewcommand{\arraystretch}{1.5}
\begin{table*}[htbp]
  \centering
  \begin{tabular}{|c|c|c|c|c|c|c|c|c|c|c|c|c|c|c|c|c|}
    \hline
    Train/Test & \multicolumn{7}{c|}{Task 1} & \multicolumn{9}{c|}{Task 2} \\
    \hline
    Class & \rot{Name} & \rot{Surname}& \rot{State}& \rot{Location}& \rot{Occupation}& \rot{$Other^{\#}$}& \rot{\textbf{$F_1$}} & \rot{Husband} & \rot{Husband's Father}& \rot{Husband's Mother}& \rot{Other Person}& \rot{Wife}& \rot{Wife's Father}& \rot{Wife's Mother}& \rot{$Other^{\#}$}& \rot{\textbf{$F_1$}} \\
    \hline
    \multicolumn{17}{|c|}{CCRF} \\
    \hline
    True/True & 0.9831 & 0.9534 & 0.9668 & 0.9546 & 0.9826 & 0.9851 & \textbf{0.9844} & 0.8089 & 0.5733 & 0.2959 & 0.6704 & 0.1992 & 0.6621 & 0.1119 & 0.9569 & \textbf{0.7333}\\
       True/Noisy & 0.9921 & 0.9742 & 0.9794 & 0.9703 & 0.9854 & 0.9887 & \textbf{0.9759} & 0.8367 & 0.653 & 0.4575 & 0.7234 & 0.2942 & 0.7014 & 0.3533 & 0.9581 & \textbf{0.7797}\\
   \hline
   
       \multicolumn{17}{|c|}{GCRF} \\
    \hline
    True/True & 0.9384 & 0.8523 & 0.8979 & 0.835 &  0.9721 & 0.9534 & \textbf{0.9218} & 0.8759 & 0.8668 & 0.912 & 0.9244 & 0.7586 & 0.8739 & 0.6793 & 0.9536 & \textbf{0.9072}\\
       True/Noisy & 0.9554 & 0.895 &  0.9174 & 0.8536 & 0.9816 & 0.9568 & \textbf{0.9345} & 0.9116 & 0.8894 & 0.9363 & 0.9479 & 0.8289 & 0.92 &  0.815 & 0.9557 & \textbf{0.9260}\\
   \hline
   
          \multicolumn{17}{|c|}{EFGCRF} \\
    \hline
    True/True & 0.9876 & 0.9611 & 0.9699 & 0.9638 & 0.9844 & 0.9868 & \textbf{0.9798} & 0.9334 & 0.922 &  0.9403 & 0.9607 & 0.9032 & 0.945 &  0.9008 & 0.9595 & \textbf{0.9430}\\
       True/Noisy & 0.9937 & 0.9758 & 0.982 &  0.9731 & 0.9864 & 0.9896 & \textbf{0.9859} & 0.9469 & 0.9341 & 0.9521 & 0.981 &  0.9352 & 0.957 & 0.9562 & 0.9623 & \textbf{0.9529}\\
   \hline
   \hline
          \multicolumn{17}{|c|}{BLSTM} \\
    \hline
    True/True & 0.9914 & 0.9667 & 0.9795 & 0.9757 & 0.987 &  0.9893 & \textbf{0.9848} & 0.9714 & 0.9864 & 0.9841 & 0.9874 & 0.9857 & 0.9832 & 0.9672 & 0.9873 & \textbf{0.9849}\\
       True/Noisy & 0.9984 & 0.9935 & 0.996 &  0.9905 & 0.9916 & 0.9964 & \textbf{0.9954} & 0.9912 & 0.9951 & 0.9958 & 0.9966 & 0.9944 & 0.994 &  0.9957 & 0.9959 & \textbf{0.9952}\\
       
   \hline
   
  \end{tabular}
  \caption{Class-wise $F_1$ scores for both semantic categories (only best performing models or models with interesting observations are shown, \# classes are not considered for IHHER evaluation)}
  \label{tab:1}
\end{table*}}
 
 \subsubsection{\textbf{Named Entity Recognition}}
We model the task of information extraction as sequential tagging and use the gold transcription to train our tagging models. However, our hypothesis is that given low CER even forced aligned HTR output can be used reliably for the task (both training and/or testing), giving reasonably close results when tested on blind test data. All our experiments result are for 10-fold cross validation on the 774 annotated samples.
%Animesh: @Herve Verify 774?
We use a character edit distance on the HTR output against the gold transcription and use the minimum CER at token boundaries to get the token level alignment for evaluating our 10-fold experiments. Hereby, we refer gold transcription as \textit{True} and the aligned HTR outputs as \textit{Noisy}. Further, we explore the inter-dependent nature of the tasks using multitask or joint learning approaches for both the semantic label spaces.

{\setlength{\tabcolsep}{3.5pt}
\renewcommand{\arraystretch}{1.5}
\begin{table}[htbp]
  \centering
  \begin{tabular}{|c|c|c|c|}
    \hline
     \multicolumn{2}{|c|}{Configuration} & \multicolumn{2}{c|}{Track}  \\
    \hline
    System & Image Segment & Basic & Complete \\
    \hline
    Naver Labs (Ours) & Line & \textbf{95.46}	& \textbf{95.03} \\
    CITlab ARGUS \cite{fornes2017icdar2017} & Line & 91.94&	91.58 \\
    CVC-UAB \cite{carbonell2018joint} & Line & 90.58 & 89.39 \\
    HMM \cite{fornes2017icdar2017} & Line & 80.28&	63.11\\
       \hline 
       \hline
   Hitsz-ICRC \cite{fornes2017icdar2017} & Word & 94.18 &	91.99 \\
    CNN \cite{fornes2017icdar2017} & Word &	79.42&	70.20\\
    \hline
  \end{tabular}
  \caption{Official IHHER evaluation, score\cite{fornes2017icdar2017} on a scale of 0-100)}
  \label{tab:2}
\end{table}}

\begin{itemize}
 
\item \textbf{Graphical Models.}
For these models, we experiment on different configurations of CRF, subsequently using a richer model or feature representation.
In all the models we use character n-gram (where n = 1, 2, 3) count (while retaining the capitalization information) as the token feature ($\sim$3K dimensions).
For each of the following models we train a system for \textit{Task 1} and \textit {Task 2} each using the \textit{True} (transcription) data, we report the trained system on the \textit{Noisy} (HTR output) and \textit{True} testing data. All the results are reported as micro $F_1$ on the 10 folds.  All the models are trained using PyStruct \cite{JMLR:v15:mueller14a} with OneSlackSSVM learner and 0.1 regularization, for 200 epochs. We train a model separately for each task. We also experiment with multitask learning. Since these model inherently computes a single hidden state assignment per observation, for joint learning we use to produce a new semantic class by taking the cross product of both semantic classes. We also experiment with NTEFGCRF \cite{meunier2017pystruct} which computes multiple hidden states per observation thereby incorporating the interaction from multiple tasks. However, none of the multitask models were able to outperform their counterpart single task models so we don't report results for those models. 

 \textbf{Chain CRF.} This model uses feature to predict labels per token  according to following equation: 
 \begin{equation*}
  \begin{aligned}
  y_i = \sum_{i\epsilon V} W^{T}_{y_{i}} \cdot node\_feature(x_i)
 \end{aligned}
 \end{equation*}
 
\textbf{Graph CRF.}
This model augments additional links (edges) to the chain model from the neighbor observations (tokens) as:
 \begin{equation*}
  \begin{aligned}
  y_i = &\sum_{i\epsilon V} W^{T}_{y_{i}} \cdot node\_feature(x_i) \\
   &+  \sum_{(i, j) \epsilon E} W^{T}_{y_{i},y_{j}} \cdot edge\_feature(x_i, x_j)
 \end{aligned}
 \end{equation*}
 The edge feature in this case is unit scalar. We experiment with link varying from length 1 to 8 (where the link of length $k$ means a connection to $k^{th}$ non-neighbor observation). The performance peaks around the length of 4-6  and therefore we pick 5 as the optimal length. The reported results have links of length 1 to 5 in addition to the chain links. The joint training strategy is the same as for chain CRF. 

 \textbf{Edge Feature Graph CRF.}
To enrich the graph CRF model further we add features on the links. We experiment with a simple feature on the link which is one hot representation of the relative distance between the current observation and connecting ($k^{th}$ non-neighbour) observation. The joint training strategy is the same as for chain CRF.

\item \textbf{Neural Models}
The state-of-the-art NER model uses BLSTM with CRF as the final layer. We experiment with and without CRF configurations.  For features, we experiment by porting the raw character n-gram counts (same as used in the graphical models) as well as using embedding. Using word embedding isn't a viable option for the given tasks since the amount of data is very limited. Further, even if a reasonable amount of data is available it's susceptible to low performance due to its high sensitivity to the performance of HTR which can exacerbate the case of OOV. 

A set of unique character can still be used to learn character embeddings (CE) from small dataset either end-to-end or via character level language modeling or sub-word embedding. When character embeddings are used we experiment with options of LSTM or CNN to compose the word embedding from the character embedding much like state-of-the-art NER model \cite{lample2016neural} but without WE. Further, when using embeddings both the character and the sequence are padded for a maximum length. All the models are implemented using Keras and have a dropout of 0.2 on all the possible parameters and trained for 10 epochs using Adam's learner. 

In our experiments, we found that even the character embedding models are not a viable option and perform extremely poor (\textless 60\% $F_1$). Further, the CRF-BLSTM surprisingly performed sub-par as compared to the BLSTM alone. We don't report results for those models.

 \textbf{BLSTM.}
In these models, we use BLSTM over the feature layer. The transformed input is further passed to a time distributed dense transform with softmax as non-linearity. The model is trained using cross entropy loss to maximize the accuracy. This model easily supports multitasking by incorporating an additional dense layer connected to the output of the LSTM. In this case, the model is trained jointly with both the labels and a loss that is average of both the individual losses. Due to the small size of the dataset we use a small model with the LSTM unit (output dimension) of size 10.  

\end{itemize}

\section{Results}
\label{section4}
From analyzing the results of various configurations \footnote{Code available at https://bitbucket.org/animeshprasad/read\_esposalles/} (10-fold experiments, Table \ref{tab:1}), we find:
\begin{itemize}
    \item Fine grain semantic class is difficult to classify (as evident from high confusion) for graphical models. This might be because of similar surface feature for two or more classes in the second label set (for example husband's and wife's father).
    \item Increasing the model complexity in graphical models while keeping the surface features from the tokens same, increases the performance. Specially so when the graphical model incorporates feature from larger context (by means of edge) which help in resolving the confusion in case of similar surface features.
    \item Due to small CER on the HTR system the models trained with the forced aligned noisy data ($\sim$2\% $F_1$ lower) as well perform near the gold data standards. 
    \item Augmenting the gold data with such noisy data doesn't improve the performance of the graphical models. The minor performance (\textless 0.005 $F_1$) increase for neural model is similar to achieved by having dropout on the feature layer.
\end{itemize}

We report our techniques performance on the blind test data as part of the IHHER official evaluation in Table \ref{tab:2}.

% {\setlength{\tabcolsep}{3.5pt}
% \renewcommand{\arraystretch}{1.5}
% \begin{table*}[t]
%   \centering
%   \begin{tabular}{|c|c|c|c|c|c|c|c|c|}
%     \hline
%      & & \multicolumn{2}{c|}{Track} & \multicolumn{5}{c|}{Basic} \\
%     \hline
%     System & \rot{Image Segment} & \rot{Basic} & \rot{Complete} & \rot{Name} & \rot{Surname}& \rot{Location}& \rot{Occupation}& \rot{State}\\
%     \hline
%     Naver Labs (Ours) & Line & 95.46	& 95.03 &	97.01 &	92.73	& 95.03 &	96.43	& 96.41 \\
%     CITlab ARGUS \cite{fornes2017icdar2017} & Line & 91.94&	91.58&	95.14&	85.78&	88.43&	93.08&	97.54 \\
%     HMM \cite{fornes2017icdar2017} & Line & 80.28&	63.11&	81.06&	60.15&	78.90&	90.23& 93.79\\
%       \hline 
%       \hline
%   Hitsz-ICRC \cite{fornes2017icdar2017} & Word & 94.18 &	91.99 &	95.68 &	91.23 &	94.93 &	93.77 &	95.35 \\
%     CNN \cite{fornes2017icdar2017} & Word &	79.42&	70.20&	83.01&	65.25&	66.31&	86.26&	97.68 \\
%     \hline
%   \end{tabular}
%   \caption{Official IHHER evaluation, score on a scale of 0-100)}
%   \label{tab:3}
% \end{table*}}

Table \ref{tab:2} shows that given high performance of character recognition standard sequential tagging models can out-perform simple RegEx as well as joint end-to-end learning models. Further, our Line segment based system outperforms even the Word segment based models. 
   
\section{Conclusion}
\label {conclusion}
For this task we use a pipeline approach where first (for HTR) the line image is preprocessed and then passed through a CNN-BLSTM architecture with CTC loss. Then (for NER) we use a  BLSTM over the feature layer (computed as character n-gram count for the tokens generated from best effort decoding of HTR output) trained using cross entropy loss to maximize the accuracy.

We find that using n-gram based features with BLSTM on IEHHR outperform end-to-end neural as well as pipeline CRF and RegEx based models. We show that for semi-structured record with near optimal HTR ($\sim$ 5\% CER), using character n-gram feature feature give approximately as good performance as NER done on ground truth.
Our analysis shows that BLSTM based models perform better than CRF based models.  Our best models shows 95.4\% score on blind test data which is $\sim$3\% better performance than the state-of-the art pattern match based techniques.

\bibliographystyle{IEEEtran}
\bibliography{main.bib}

\end{document}